\documentclass[sigconf]{acmart}

\AtBeginDocument{%
  \providecommand\BibTeX{{%
    \normalfont B\kern-0.5em{\scshape i\kern-0.25em b}\kern-0.8em\TeX}}}

\setcopyright{acmcopyright}
\copyrightyear{2023}
\acmYear{2023}
\acmDOI{XXXXXXX.XXXXXXX}

\acmConference[]{RLEM '23}{November 11--12,
  2023}{Istanbul, Turkey}

\begin{document}

\title{A Lightweight Calibrated Simulation Enabling Efficient Offline Learning for Optimal Control of Real Buildings}


\author{Judah Goldfeder}
\affiliation{%
  \institution{Google, Columbia University}
  \country{United States}
}
\email{judahg@google.com}

\author{John Sipple}
\affiliation{%
  \institution{Google}
  \country{United States}
}
\email{sipple@google.com}

\renewcommand{\shortauthors}{Goldfeder and Sipple}

\begin{abstract}
Modern commercial Heating, Ventilation, and Air Conditioning (HVAC) devices form a complex and interconnected thermodynamic system with the building and outside weather conditions, and current setpoint control policies are not fully optimized for minimizing energy use and carbon emission. Given a suitable training environment, a Reinforcement Learning (RL) model is able to improve upon these policies, but training such a model, especially in a way that scales to thousands of buildings, presents many real world challenges. We propose a novel simulation-based approach, where a customized simulator is used to train the agent for each building. Our open-source simulator\footnote{Available online: https://github.com/google/sbsim} is lightweight and calibrated via telemetry from the building to reach a higher level of fidelity. On a two-story, 68,000 square foot building, with 127 devices, we were able to calibrate our simulator to have just over half a degree of drift from the real world over a six-hour interval. This approach is an important step toward having a real-world RL control system that can be scaled to many buildings, allowing for greater efficiency and resulting in reduced energy consumption and carbon emissions. 

\end{abstract}



\maketitle

\section{Introduction}
Energy optimization and management in commercial buildings is a very important problem, whose importance is only growing with time. Buildings account for 39\% of all US carbon emissions \cite{lu2020review}. Reducing those emissions by even a small percentage can have a significant
effect, especially in more extreme climates.
Our contributions include a highly customizable and scalable HVAC and building simulator, a rapid configuration method to customize the simulator to a particular building, a calibration method to improve this fidelity using real world data, and an evaluation method to measure the simulator fidelity. 


\textbf{Overview of the HVAC problem}
Most office buildings are equipped with advanced HVAC devices, like Variable Air Volume (VAV) devices, Hot Water Systems (HWS), Air Conditioner (AC) and Air Handlers that are configured and tuned by the engineers, manufacturers, installers, and operators to run efficiently with the device’s local control loops \cite{mcquiston2023heating}. However, integrating multiple HVAC devices from diverse vendors into a building “system” requires technicians to program fixed operating conditions for these units, which may not be optimal for every building and every potential weather condition. If existing setpoint control policies are not optimal under all conditions, the possibility exists that an ML model may be trained to continuously tune a small number of setpoints to achieve greater energy efficiency and reduced carbon emission \cite{wei2017deep}.  

\textbf{RL for HVAC systems} We define the state of the office building  $S_t$  at time $t$ as a fixed length vector of measurements from sensors on the building's devices, such as a specific VAV’s zone air temperature, gas meter’s flow rate, etc. The action on the building $A_t$ is a fixed-length vector of device setpoints selected by the agent at time $t$, such as the boiler's supply water temperature setpoint, etc. We also define a custom feedback signal, or reward,  that indicates the quality of taking an action in the current state, as a weighted sum of negative cost functions for carbon emission, energy consumption, and zone-level setpoint deviation. The goal of RL is to maximize the expected long-term cumulative reward \cite{sutton2018reinforcement}. In order to converge to the optimal policy, the agent requires many training iterations, making training directly on the building from scratch inefficient and impracticable. Therefore, it is necessary to train in an offline sandbox environment that adequately emulates the dynamics of the building before being deployed on the actual building. 

\section{Related Work}
Considerable attention has been paid to HVAC control in recent years, and a growing portion of that has considered how RL and its various associated algorithms can be leveraged \cite{gao2023comparative,mason2019review,wang2023comparison,yu2021review,yu2020multi}. A central requirement in RL is the offline environment that trains the RL agent. Several methods for this have been proposed, largely falling under three broad categories.

\textbf{Data-driven Emulators}: Some works attempt to learn a dynamics model as a multivariate regression model from real world data \cite{zhang2019building,zou2020towards}, often using recurrent neural network architectures, such as LSTMs \cite{sendra2020long,velswamy2017long,,zhuang2023data}. The difficulty here is that data-driven models often do not generalize well to circumstances outside the training distribution, especially since they are not physics based.

\textbf{Physics-based Simulation}:  EnergyPlus \cite{crawley2001energyplus}, a high-fidelity simulator developed by the Department of Energy, is commonly used \cite{azuatalam2020reinforcement,wani2019control,wei2017deep,zhao2015energyplus}, but suffers from the scalability issues outlined above.

\textbf{Offline RL}: The second approach is to train the agent directly from the historical real world data, without ever producing an interactive environment \cite{blad2022data,chen2020gnu,chen2023fast}. While the real world data is obviously of high accuracy and quality, this presents a major challenge, since the agent cannot take actions in the real world and interact with any form of an environment, severely limiting its ability to improve over the baseline policy that produced the real world data \cite{levine2020offline}.

To overcome the limitations of each of the above three methods, some work has proposed a hybrid approach \cite{balali2023energy,zhao2021hybrid}, and this is the category our work falls under. Our approach uses a physics-based simulator that achieves an ideal balance between speed and fidelity, which is sufficient to train an effective control agent off-line.

\section{A simple calibrated Simulation}
\textbf{Design Considerations}
A fundamental tradeoff when designing a simulator is speed versus fidelity. Fidelity is the simulator’s ability to reproduce the building’s true dynamics that affect the optimization process. Speed refers to both simulator configuration time, and agent training time, i.e., the time necessary for the agent to optimize its policy using the simulator.
Every building is unique, due to its physical layout, equipment, and location. Fully customizing a high fidelity simulation to a specific target building requires nearly exhaustive knowledge of the building structure, materials, location, etc., some of which is unknowable. Also, the configuration time required for high-fidelity simulations limits their utility for deploying RL-based optimization to many buildings.
Our goal is to develop a method for applying RL at scale to commercial buildings. To this end, we  must have a simulated environment to train the agent, as real world training is not possible.
In order to scale to many buildings it must be easy for our simulator to be configured to a new building, with enough fidelity to the real world to be useful. We note that while RL is the control method proposed in this work, our calibrated simulator is useful as well for other controller methods such as Model Predictive Control \cite{kouvaritakis2016model}.
To meet these requirements we designed a lightweight simulator based on Finite Differences approximation of heat exchange, with a novel automated procedure to go from building floor plans to a custom simulator with little manual effort. For improved fidelity, we designed a calibration and evaluation pipeline based on real telemetry.

\textbf{Building Thermal Model}
We developed our simulation around our abstract thermal model for office buildings, shown in Figure 1.  A building consists of conditioned zones, where the measured zone temperature,  $T_z$, should be within upper and lower setpoints, $\hat{T}_{z, max}$ and $\hat{T}_{z,min}$. Thermal power for heating or cooling is supplied to each zone, $\dot{Q}_s$, and recirculated from the zone, $\dot{Q}_r$ from the HVAC system, with additional thermal exchange $\dot{Q}_z$ from walls, doors, etc. The Air Handler supplies the building with air at supply air temperature setpoint $\hat{T}_s$ drawing fresh air, $\dot{m}_{amb}$ at ambient temperatures, $T_{amb}$ and returning exhaust air $\dot{m}_{exhaust}$ at temperature $T_{exhaust}$ to the outside using intake and exhaust fans: $\dot{W}_{a,in}$ and $\dot{W}_{a,out}$.  Some of the return air can be recirculated, $\dot{m}_{recirc}$. Central air conditioning is achieved with a chiller and pump that joins a refrigeration cycle to the supply air, consuming electrical energy for the AC compressor $\dot{W}_{c}$ and coolant circulation,  $\dot{W}_{c,p}$. The hot water cycle consists of a boiler that maintains the supply water temperature at $T_b$ heated by natural gas power  $\dot{Q}_{b}$, and a pump that circulates hot water through the building, with electrical power  $\dot{W}_{b,p}$.  Supply air is delivered to the zones through VAV devices.






We selected \textbf{water supply temperature} $\hat{T}_b$ and the \textbf{air handler supply temperature}  $\hat{T}_s$ as agent actions because they affect the balance of electricity and natural gas consumption, they affect multiple device interactions, and they affect occupant comfort.


\textbf{Finite Differences Approximation}
The diffusion of thermal energy in time and space of the building can be approximated using the method of Finite Differences (FD)\cite{Sparrow,lomax2002fundamentals}, and applying an energy balance.  This method divides each floor of the building into a grid of three-dimensional control volumes and applies thermal diffusion equations to estimate the temperature of each control volume. By assuming each floor is thermally isolated, (i.e., no heat is transferred between floors), we can simplify the three-spatial dimensions into a spatial two-dimensional heat transfer problem.  Each control volume is a narrow volume bounded horizontally, parameterized by $\Delta x^2$, and vertically by the height of the floor. 
The energy balance, shown below, is applied to each discrete control volume in the FD grid, and consists of the following components: (a) the thermal exchange across each face of the four participating faces control volume via conduction or convection $Q_1$, $Q_2$, $Q_3$, $Q_4$, (b) the change in internal energy over time in the control volume $Mc \frac{\Delta T}{\Delta t}$, and (c) an external energy source that enables applying local thermal energy from the HVAC model only for those control volumes that include an airflow diffuser, $Q_{ext}$. The equation is $Q_{ext}+Q_1+Q_2+Q_3+Q_4=Mc \frac{\Delta T}{\Delta t}$, where $M$ is the mass and $c$ is the heat capacity of the control volume,  $\Delta T$ is the temperature change from the prior timestep and $\Delta t$ is the timestep interval. 

\begin{figure}[H]
\includegraphics[width=8cm]{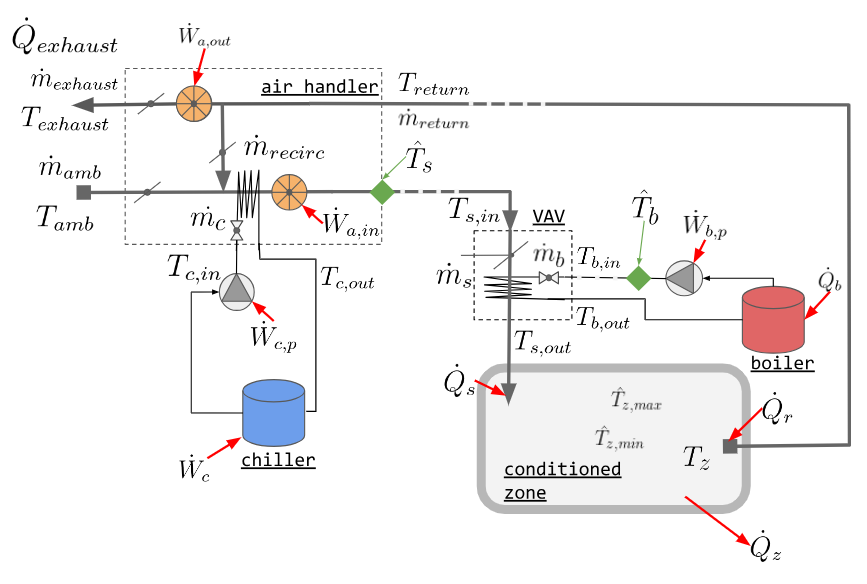}
\centering
\caption{ Building Thermal Model 
}
\end{figure}

The thermal exchange in (a) is calculated using Fourier’s law of steady conduction in the interior control volumes (walls and interior air), parameterized by the conductivity of the volume, and the exchange across the exterior faces of control volumes is calculated using the forced convection equation, parameterized by the convection coefficient, which approximates winds and currents surrounding the building. The change in internal energy (b) is parameterized by the density, and heat capacity of the control volume. Finally, the thermal energy associated with the VAV (c) is equally distributed to all associated control volumes that have a diffuser. 


Thermal diffusion within the building is mainly accomplished via forced or natural convection currents, which can be notoriously difficult to estimate accurately. We note that heat transfer using air circulation is effectively the exchange of air mass between control volumes, which we approximate by a randomized shuffling of air within zones, parameterized by a shuffle probability.

\textbf{HVAC Model}
The HVAC system is modeled as an energy balance with a lossy thermal exchange between the hot water and air circuits, evaluating temperature differences, and air and water flow rates with efficiency parameters. Aggregate heating and cooling demands at the AC and HWS are governed by thermostats that create demand when its zone temperature is near zone air temperature setpoints. Special control volumes associated with air diffusers allow thermal energy exchange between the HVAC model and the FD grid, $\dot{Q}_s$.

\textbf{Simulator Configuration}
For RL to scale to many buildings, it is critical to be able to easily and rapidly configure the simulator to any arbitrary building. To configure the simulator, we require floorplans, and HVAC metadata.
We preprocess the detailed floorplan blueprints of the building, and extract a grid that gives us an approximate placement of walls. We also employ a user interface to label the location of each HVAC device on the floorplan grid. We tested this pipeline on our pilot building, which consisted of two floors with combined surface area of 68,000 $\text{ft}^2$, and 127 HVAC devices. Given floorplans and HVAC layout information, a single technician was able to generate a fully specified simulation in under 3 hours that matched the real building in every device and room. 

\textbf{Simulator Calibration and Evaluation}
In order to calibrate the simulator to the real world using data, we must have a metric with which to evaluate our simulator, and an optimization method to improve our simulator on this metric. 
We proposed a novel evaluation procedure, based on $N$-step prediction. Each iteration step of our simulator was designed to represent a five minute interval, and our real world data is also sampled in five minute intervals. 
To evaluate the simulator, we take a chunk of real data of $N$ observations. We then initialize the simulator so that its initial state matches that of the starting observation, and run the simulator for $N$ steps, replaying the same HVAC policy as was used in the real world observations. Next, we calculate our simulation fidelity metric, which is the mean absolute error of the measured zone air temperatures, at the $N$th timestep. 
More formally, we define the spatial Mean Absolute Error (MAE) of $Z$ zones at timestep $t$ as $\epsilon_t = \frac{1}{Z} \sum^{Z}_{z=1} | T_{real,t,z} -  T_{sim,t,z}|$, where $T_{real,t,z}$ is the measured zone air temperature for zone $z$ at timestamp $t$, and $ T_{sim,t,z} = \frac{1}{C_z} \sum^{C_z}_{c=1} T_{t,c} $ is the mean temperature of all control volumes $C_z$ in zone $z$ at time $t$.
Thus, to evaluate the simulator on $N$-step prediction, we run the simulator for timesteps $0$ to $N-1$, and calculate the above metric for $t = N-1$.

\textbf{Simulator Calibration}
Once we defined our simulation fidelity metric, we can minimize the error by searching over the physical parameters that affect the simulation response.
The variables tuned during the parameter search included the following: (a) the forced convection coefficient quantifying outside wind and air currents against the building exterior surfaces; (b) the thermal conductivity, heat capacity, and density of exterior and interior walls; and (c) the shuffle probability that controls how likely an air volume will be exchanged with another within the zone to approximate the internal air circulation and interior forced convection.

\section{Experiment Results}
We now demonstrate the results of how our simulator, when tuned and calibrated, is able to make real-world predictions.

\textbf{Experiment Setup}
To test out our simulator, we obtained telemetry recordings from our pilot building, a commercial office building located in Mountain View, California. The building has two stories with a combined surface area of 68,000 square feet, and has 127 HVAC devices. We obtained floor plan blueprints and used them to configure a customized simulator for the building, a process that took a single human less than three hours to complete.

\textbf{Calibration Data}
To tune and evaluate our simulator, we took three, non-overlapping intervals of telemetry: (a) Thursday, July 6 2023, from 1:40 AM to 7:40 AM ($N$ = 72), (b) Friday, July 7 2023, from 1:40 AM to 7:40 AM ($N$ = 72), and (c) from Friday, July 7 2023, from 11:40 AM to 5:40 PM ($N$ = 72). The first telemetry interval, (a), was used to tune the simulator for calibration. The second and third telemetry intervals, (b) and (c), were used for validation.
\begin{table}[H]
  \caption{Hyperparameter ranges and chosen values}
  \label{tab:freq}
  \begin{tabular}{cccl}
    \toprule
    Hyperparameter&range&best\\
    \midrule
    exterior convection coefficient ($W/m^2/K$) & 5 - 800 & 800 \\
    exterior wall conductivity ($W/m/K$) & 0.01 - 1 & 0.01 \\
    exterior wall density ($kg/m^3$) & 1 - 3000 & 2748 \\
    exterior wall heat capacity $(J/Kg/K)$ & 100 - 2500 & 2500 \\
    interior wall conductivity ($W/m/K$) & 5 - 800 & 780 \\
    interior wall density ($kg/m^3$) & 0.5 - 1500 & 0.5 \\
    interior wall heat capacity $(J/Kg/K)$ & 500 - 1500 & 500 \\
    interior air volume shuffle probability  & 0 - 1 & 1 \\
  \bottomrule
\end{tabular}
\end{table}
\textbf{Calibration Procedure}
We tuned for 100 iterations on our simulator. The parameters varied, the sweep range limits, and the values found that minimized the calibration metric are shown in Table 1. Because of our focus on speed, we can similate a five minute interval in about 15 seconds on one CPU, allowing for very rapid tuning.
We reviewed the parameters that yielded the lowest simulation error from calibration. Densities, heat capacities, and conductivities plausibly matched common interior and exterior building materials. However, the external convection coefficient was higher than under normal weather conditions, and likely is compensating for radiative losses and gains, which were not directly simulated.

\textbf{Calibration Results}
Table 2 shows the $N$-step prediction fidelity, as MAE, over a six-hour prediction window ($N=72$). We calculated the spatial mean absolute temperature error, as defined above. We also present a second metric, the median spatial temperature error. This was not used in the tuning process, but gives us some insight into how well the calibration process is performing.
As indicated in Table 2, our tuning procedure drifts only $0.6^\circ$  on average over a 6 hour period on the tuning set, and we get good generalization: almost identical error for the same time period of a different day. It should be noted that an uncalibrated model had a much larger error of $2.1^\circ$.
Furthermore, the median temperature in these two cases is near zero, indicating that even despite drift, our prediction remains a mostly unbiased estimator.
For the different time of day scenario, we observe a larger drift of $1.2^\circ$, and the median is shifted as well, indicated that tuning on one time of day is not as useful when making predictions on another.

\begin{table}[H]
  \caption{Mean and Median Absolute Error on $N=72$ prediction window}
  \label{tab:freq}
  \begin{tabular}{c|ccl}
    \toprule
    Metric&Tuning Data (a) & Validation (b) & Validation (c)\\
    \midrule

    MAE &  0.64 $^\circ C$ & 0.63 $^\circ C$ & 1.18 $^\circ C$\\
    Median & 0.01 $^\circ C$& 0.18 $^\circ C$& 0.98 $^\circ C$ \\

  \bottomrule
\end{tabular}
\end{table}

\begin{figure}[H]
\includegraphics[width=8cm]{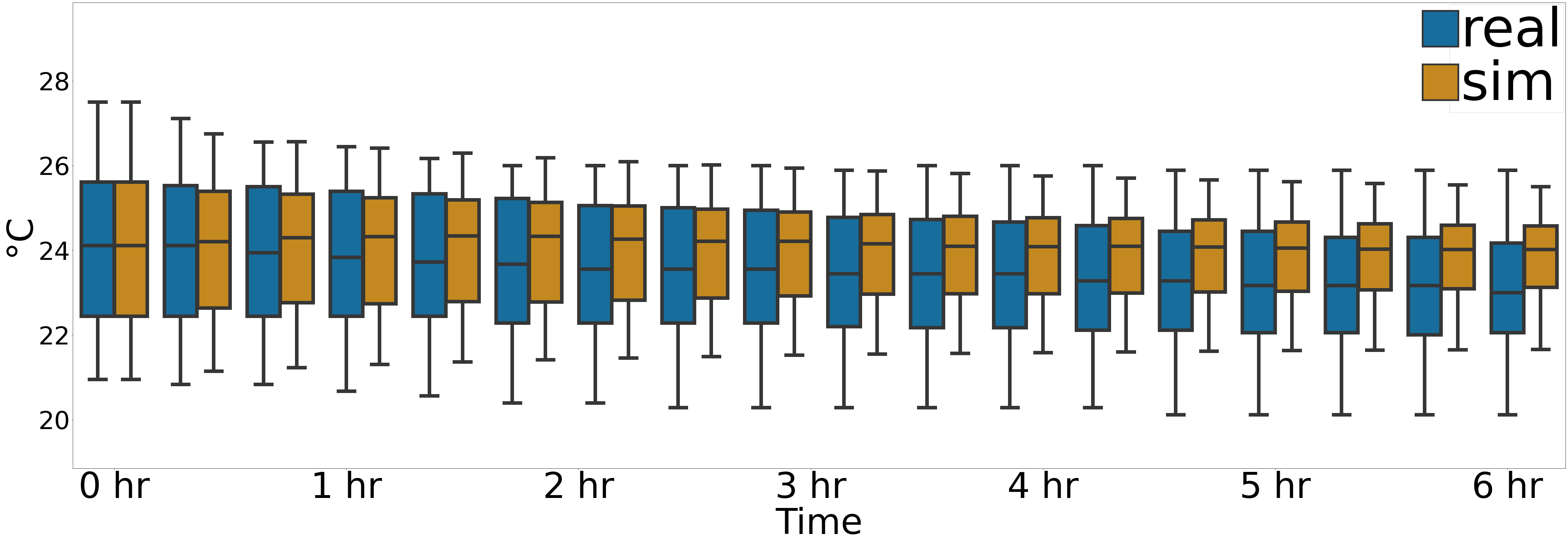}
\centering
\caption{ Temperature Drift Over 6 hours, from validation (b).}
\end{figure}

\textbf{Visualizing Temperature Drift Over Time}
Figure 2 illustrates temperature drift over time. At each time step, we
plot the real world temperatures in a blue boxplot and the simulator ones in orange.

\textbf{Visualizing Spatial Errors}
Figure 3 is a heatmap of the spatial temperature difference across both floors after six hours. Red regions indicate the simulator was warmer than the validation data, blue regions indicate cooler, and white regions had no temperature difference. 
The ring of blue around the building indicates that our simulator is too cold on the perimeter, which implies that the heat exchange with the outside is happening more rapidly than it would in the real world. The inside of the building remains red, which means that despite the simulator perimeter being cooler than the real world, the inside is warmer. 
The white band region running along the perimeter is where the temperature achieved a perfect fit with the validation data. The interior red regions are due to slight errors in modeling interior heat exchange, and the blue regions indicate errors due to high thermal losses with the outside. We suspect these errors are largely due to assuming thermal isolation between floors and assuming no radiative exchange with the outside.  Consequently, the calibration process compensates with a high convection coefficient. Internal thermal exchange seems slower than the real world, and the calibration compensates with more rapid outside thermal exchange.
\begin{figure}[H]
\includegraphics[width=8cm]{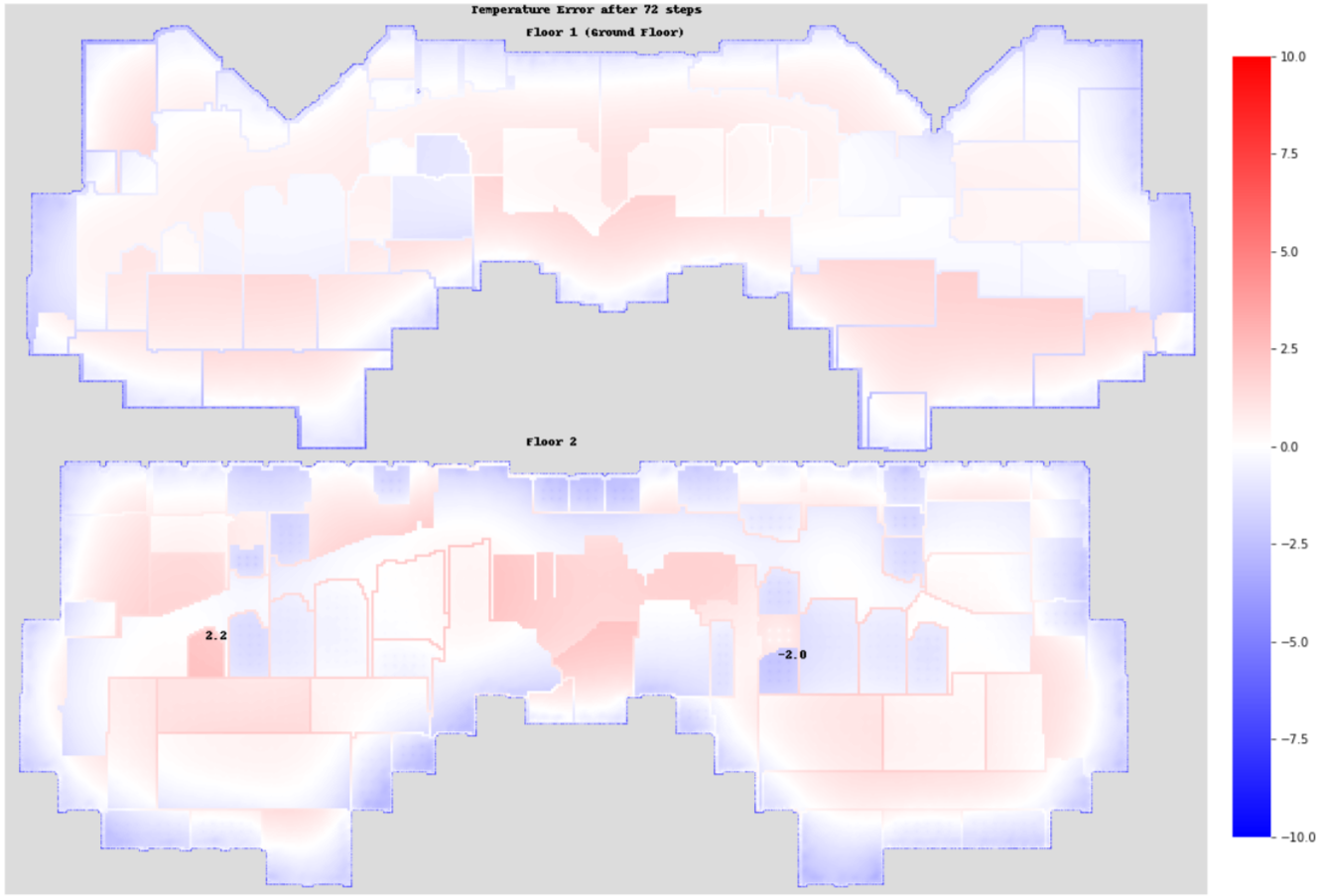}
\centering
\caption{Heatmap representing the temperature difference between the simulator and the real world after 6 hours, with red indicating the simulator is hotter, and blue colder.}
\end{figure}

\section{Discussion and Future Work}

Developing a simulator is an important step toward achieving our objective of training and deploying an RL agent on the actual building.
We already have a real building that we have the ability to control, and using our pipeline produced a calibrated simulator that perfectly matches all of its devices, setpoints and sensors. 
Our future goal, showing how an agent trained on this simulator is able to produce a useful RL policy on the real building, will require a few more components. In future live trials on the real building, we will train an RL agent on our simulator, and measure its performance in the real world. Care must be taken in this process, as RL agents are prone to learning false patterns from the simulator \cite{arroyo2022reinforced,liu2006experimental}. 

We also have outlined several approaches that we intend to use to further improve our simulator and fine tuning process
\begin{enumerate}
\item Tuning on data from multiple seasons, various times of day, and longer time windows, to improve generalizability. 

\item Updating our simulation by adding in a radiative heat transfer model that enables solar gains during the daytime and losses at night.

\item Expanding the action space to include AC static pressure and HWS differential pressure that govern the air and water flow rates. 

\item Adding a data-driven predictor-corrector to improve fidelity.

\end{enumerate}
We are optimistic that our novel simulation tuning process will allow us to develop a useful and transferable, HVAC control solution, and we hope our work will inspire further effort in this field.



\citestyle{acmnumeric}  

\bibliographystyle{ACM-Reference-Format}
\bibliography{buildsys23-43}


\begin{thebibliography}{29}


\ifx \showCODEN    \undefined \def \showCODEN     #1{\unskip}     \fi
\ifx \showDOI      \undefined \def \showDOI       #1{#1}\fi
\ifx \showISBNx    \undefined \def \showISBNx     #1{\unskip}     \fi
\ifx \showISBNxiii \undefined \def \showISBNxiii  #1{\unskip}     \fi
\ifx \showISSN     \undefined \def \showISSN      #1{\unskip}     \fi
\ifx \showLCCN     \undefined \def \showLCCN      #1{\unskip}     \fi
\ifx \shownote     \undefined \def \shownote      #1{#1}          \fi
\ifx \showarticletitle \undefined \def \showarticletitle #1{#1}   \fi
\ifx \showURL      \undefined \def \showURL       {\relax}        \fi
\providecommand\bibfield[2]{#2}
\providecommand\bibinfo[2]{#2}
\providecommand\natexlab[1]{#1}
\providecommand\showeprint[2][]{arXiv:#2}

\bibitem[Arroyo et~al\mbox{.}(2022)]%
        {arroyo2022reinforced}
\bibfield{author}{\bibinfo{person}{Javier Arroyo}, \bibinfo{person}{Carlo
  Manna}, \bibinfo{person}{Fred Spiessens}, {and} \bibinfo{person}{Lieve
  Helsen}.} \bibinfo{year}{2022}\natexlab{}.
\newblock \showarticletitle{Reinforced model predictive control (RL-MPC) for
  building energy management}.
\newblock \bibinfo{journal}{\emph{Applied Energy}}  \bibinfo{volume}{309}
  (\bibinfo{year}{2022}), \bibinfo{pages}{118346}.
\newblock


\bibitem[Azuatalam et~al\mbox{.}(2020)]%
        {azuatalam2020reinforcement}
\bibfield{author}{\bibinfo{person}{Donald Azuatalam}, \bibinfo{person}{Wee-Lih
  Lee}, \bibinfo{person}{Frits de Nijs}, {and} \bibinfo{person}{Ariel
  Liebman}.} \bibinfo{year}{2020}\natexlab{}.
\newblock \showarticletitle{Reinforcement learning for whole-building HVAC
  control and demand response}.
\newblock \bibinfo{journal}{\emph{Energy and AI}}  \bibinfo{volume}{2}
  (\bibinfo{year}{2020}), \bibinfo{pages}{100020}.
\newblock


\bibitem[Balali et~al\mbox{.}(2023)]%
        {balali2023energy}
\bibfield{author}{\bibinfo{person}{Yasaman Balali}, \bibinfo{person}{Adrian
  Chong}, \bibinfo{person}{Andrew Busch}, {and} \bibinfo{person}{Steven
  O’Keefe}.} \bibinfo{year}{2023}\natexlab{}.
\newblock \showarticletitle{Energy modelling and control of building heating
  and cooling systems with data-driven and hybrid models—A review}.
\newblock \bibinfo{journal}{\emph{Renewable and Sustainable Energy Reviews}}
  \bibinfo{volume}{183} (\bibinfo{year}{2023}), \bibinfo{pages}{113496}.
\newblock


\bibitem[Blad et~al\mbox{.}(2022)]%
        {blad2022data}
\bibfield{author}{\bibinfo{person}{Christian Blad}, \bibinfo{person}{Simon
  B{\o}gh}, {and} \bibinfo{person}{Carsten~Skovmose Kalles{\o}e}.}
  \bibinfo{year}{2022}\natexlab{}.
\newblock \showarticletitle{Data-driven offline reinforcement learning for
  HVAC-systems}.
\newblock \bibinfo{journal}{\emph{Energy}}  \bibinfo{volume}{261}
  (\bibinfo{year}{2022}), \bibinfo{pages}{125290}.
\newblock


\bibitem[Chen et~al\mbox{.}(2020)]%
        {chen2020gnu}
\bibfield{author}{\bibinfo{person}{Bingqing Chen}, \bibinfo{person}{Zicheng
  Cai}, {and} \bibinfo{person}{Mario Berg{\'e}s}.}
  \bibinfo{year}{2020}\natexlab{}.
\newblock \showarticletitle{Gnu-rl: A practical and scalable reinforcement
  learning solution for building hvac control using a differentiable mpc
  policy}.
\newblock \bibinfo{journal}{\emph{Frontiers in Built Environment}}
  \bibinfo{volume}{6} (\bibinfo{year}{2020}), \bibinfo{pages}{562239}.
\newblock


\bibitem[Chen et~al\mbox{.}(2023)]%
        {chen2023fast}
\bibfield{author}{\bibinfo{person}{Liangliang Chen}, \bibinfo{person}{Fei
  Meng}, {and} \bibinfo{person}{Ying Zhang}.} \bibinfo{year}{2023}\natexlab{}.
\newblock \showarticletitle{Fast Human-in-the-loop Control for HVAC Systems via
  Meta-learning and Model-based Offline Reinforcement Learning}.
\newblock \bibinfo{journal}{\emph{IEEE Transactions on Sustainable Computing}}
  (\bibinfo{year}{2023}).
\newblock


\bibitem[Crawley et~al\mbox{.}(2001)]%
        {crawley2001energyplus}
\bibfield{author}{\bibinfo{person}{Drury~B Crawley}, \bibinfo{person}{Linda~K
  Lawrie}, \bibinfo{person}{Frederick~C Winkelmann}, \bibinfo{person}{Walter~F
  Buhl}, \bibinfo{person}{Y~Joe Huang}, \bibinfo{person}{Curtis~O Pedersen},
  \bibinfo{person}{Richard~K Strand}, \bibinfo{person}{Richard~J Liesen},
  \bibinfo{person}{Daniel~E Fisher}, \bibinfo{person}{Michael~J Witte},
  {et~al\mbox{.}}} \bibinfo{year}{2001}\natexlab{}.
\newblock \showarticletitle{EnergyPlus: creating a new-generation building
  energy simulation program}.
\newblock \bibinfo{journal}{\emph{Energy and buildings}} \bibinfo{volume}{33},
  \bibinfo{number}{4} (\bibinfo{year}{2001}), \bibinfo{pages}{319--331}.
\newblock


\bibitem[Gao and Wang(2023)]%
        {gao2023comparative}
\bibfield{author}{\bibinfo{person}{Cheng Gao} {and} \bibinfo{person}{Dan
  Wang}.} \bibinfo{year}{2023}\natexlab{}.
\newblock \showarticletitle{Comparative study of model-based and model-free
  reinforcement learning control performance in HVAC systems}.
\newblock \bibinfo{journal}{\emph{Journal of Building Engineering}}
  \bibinfo{volume}{74} (\bibinfo{year}{2023}), \bibinfo{pages}{106852}.
\newblock


\bibitem[Kouvaritakis and Cannon(2016)]%
        {kouvaritakis2016model}
\bibfield{author}{\bibinfo{person}{Basil Kouvaritakis} {and}
  \bibinfo{person}{Mark Cannon}.} \bibinfo{year}{2016}\natexlab{}.
\newblock \showarticletitle{Model predictive control}.
\newblock \bibinfo{journal}{\emph{Switzerland: Springer International
  Publishing}}  \bibinfo{volume}{38} (\bibinfo{year}{2016}).
\newblock


\bibitem[Levine et~al\mbox{.}(2020)]%
        {levine2020offline}
\bibfield{author}{\bibinfo{person}{Sergey Levine}, \bibinfo{person}{Aviral
  Kumar}, \bibinfo{person}{George Tucker}, {and} \bibinfo{person}{Justin Fu}.}
  \bibinfo{year}{2020}\natexlab{}.
\newblock \showarticletitle{Offline reinforcement learning: Tutorial, review,
  and perspectives on open problems}.
\newblock \bibinfo{journal}{\emph{arXiv preprint arXiv:2005.01643}}
  (\bibinfo{year}{2020}).
\newblock


\bibitem[Liu and Henze(2006)]%
        {liu2006experimental}
\bibfield{author}{\bibinfo{person}{Simeng Liu} {and} \bibinfo{person}{Gregor~P
  Henze}.} \bibinfo{year}{2006}\natexlab{}.
\newblock \showarticletitle{Experimental analysis of simulated reinforcement
  learning control for active and passive building thermal storage inventory:
  Part 2: Results and analysis}.
\newblock \bibinfo{journal}{\emph{Energy and buildings}} \bibinfo{volume}{38},
  \bibinfo{number}{2} (\bibinfo{year}{2006}), \bibinfo{pages}{148--161}.
\newblock


\bibitem[Lomax et~al\mbox{.}(2002)]%
        {lomax2002fundamentals}
\bibfield{author}{\bibinfo{person}{Harvard Lomax}, \bibinfo{person}{Thomas~H
  Pulliam}, \bibinfo{person}{David~W Zingg}, {and} \bibinfo{person}{TA
  Kowalewski}.} \bibinfo{year}{2002}\natexlab{}.
\newblock \showarticletitle{Fundamentals of computational fluid dynamics}.
\newblock \bibinfo{journal}{\emph{Appl. Mech. Rev.}} \bibinfo{volume}{55},
  \bibinfo{number}{4} (\bibinfo{year}{2002}), \bibinfo{pages}{B61--B61}.
\newblock


\bibitem[Lu and Lai(2020)]%
        {lu2020review}
\bibfield{author}{\bibinfo{person}{Mengxue Lu} {and} \bibinfo{person}{Joseph
  Lai}.} \bibinfo{year}{2020}\natexlab{}.
\newblock \showarticletitle{Review on carbon emissions of commercial
  buildings}.
\newblock \bibinfo{journal}{\emph{Renewable and Sustainable Energy Reviews}}
  \bibinfo{volume}{119} (\bibinfo{year}{2020}), \bibinfo{pages}{109545}.
\newblock


\bibitem[Mason and Grijalva(2019)]%
        {mason2019review}
\bibfield{author}{\bibinfo{person}{Karl Mason} {and} \bibinfo{person}{Santiago
  Grijalva}.} \bibinfo{year}{2019}\natexlab{}.
\newblock \showarticletitle{A review of reinforcement learning for autonomous
  building energy management}.
\newblock \bibinfo{journal}{\emph{Computers \& Electrical Engineering}}
  \bibinfo{volume}{78} (\bibinfo{year}{2019}), \bibinfo{pages}{300--312}.
\newblock


\bibitem[McQuiston et~al\mbox{.}(2023)]%
        {mcquiston2023heating}
\bibfield{author}{\bibinfo{person}{Faye~C McQuiston}, \bibinfo{person}{Jerald~D
  Parker}, \bibinfo{person}{Jeffrey~D Spitler}, {and} \bibinfo{person}{Hessam
  Taherian}.} \bibinfo{year}{2023}\natexlab{}.
\newblock \bibinfo{booktitle}{\emph{Heating, ventilating, and air conditioning:
  analysis and design}}.
\newblock \bibinfo{publisher}{John Wiley \& Sons}.
\newblock


\bibitem[Sendra-Arranz and Guti{\'e}rrez(2020)]%
        {sendra2020long}
\bibfield{author}{\bibinfo{person}{R Sendra-Arranz} {and} \bibinfo{person}{A
  Guti{\'e}rrez}.} \bibinfo{year}{2020}\natexlab{}.
\newblock \showarticletitle{A long short-term memory artificial neural network
  to predict daily HVAC consumption in buildings}.
\newblock \bibinfo{journal}{\emph{Energy and Buildings}}  \bibinfo{volume}{216}
  (\bibinfo{year}{2020}), \bibinfo{pages}{109952}.
\newblock


\bibitem[Sparrow(1993)]%
        {Sparrow}
\bibfield{author}{\bibinfo{person}{E~M Sparrow}.}
  \bibinfo{year}{1993}\natexlab{}.
\newblock \bibinfo{title}{Heat Transfer: Conduction [Lecture Notes]}.
\newblock
\newblock


\bibitem[Sutton and Barto(2018)]%
        {sutton2018reinforcement}
\bibfield{author}{\bibinfo{person}{Richard~S Sutton} {and}
  \bibinfo{person}{Andrew~G Barto}.} \bibinfo{year}{2018}\natexlab{}.
\newblock \bibinfo{booktitle}{\emph{Reinforcement learning: An introduction}}.
\newblock \bibinfo{publisher}{MIT press}.
\newblock


\bibitem[Velswamy et~al\mbox{.}(2017)]%
        {velswamy2017long}
\bibfield{author}{\bibinfo{person}{Kirubakaran Velswamy}, \bibinfo{person}{Biao
  Huang}, {et~al\mbox{.}}} \bibinfo{year}{2017}\natexlab{}.
\newblock \showarticletitle{A Long-Short Term Memory Recurrent Neural Network
  Based Reinforcement Learning Controller for Office Heating Ventilation and
  Air Conditioning Systems}.
\newblock  (\bibinfo{year}{2017}).
\newblock


\bibitem[Wang et~al\mbox{.}(2023)]%
        {wang2023comparison}
\bibfield{author}{\bibinfo{person}{Marshall Wang}, \bibinfo{person}{John
  Willes}, \bibinfo{person}{Thomas Jiralerspong}, {and} \bibinfo{person}{Matin
  Moezzi}.} \bibinfo{year}{2023}\natexlab{}.
\newblock \showarticletitle{A Comparison of Classical and Deep Reinforcement
  Learning Methods for HVAC Control}.
\newblock \bibinfo{journal}{\emph{arXiv preprint arXiv:2308.05711}}
  (\bibinfo{year}{2023}).
\newblock


\bibitem[Wani et~al\mbox{.}(2019)]%
        {wani2019control}
\bibfield{author}{\bibinfo{person}{Mubashir Wani}, \bibinfo{person}{Akshya
  Swain}, {and} \bibinfo{person}{Abhisek Ukil}.}
  \bibinfo{year}{2019}\natexlab{}.
\newblock \showarticletitle{Control strategies for energy optimization of HVAC
  systems in small office buildings using energyplus tm}. In
  \bibinfo{booktitle}{\emph{2019 IEEE Innovative Smart Grid Technologies-Asia
  (ISGT Asia)}}. IEEE, \bibinfo{pages}{2698--2703}.
\newblock


\bibitem[Wei et~al\mbox{.}(2017)]%
        {wei2017deep}
\bibfield{author}{\bibinfo{person}{Tianshu Wei}, \bibinfo{person}{Yanzhi Wang},
  {and} \bibinfo{person}{Qi Zhu}.} \bibinfo{year}{2017}\natexlab{}.
\newblock \showarticletitle{Deep reinforcement learning for building HVAC
  control}. In \bibinfo{booktitle}{\emph{Proceedings of the 54th annual design
  automation conference 2017}}. \bibinfo{pages}{1--6}.
\newblock


\bibitem[Yu et~al\mbox{.}(2021)]%
        {yu2021review}
\bibfield{author}{\bibinfo{person}{Liang Yu}, \bibinfo{person}{Shuqi Qin},
  \bibinfo{person}{Meng Zhang}, \bibinfo{person}{Chao Shen},
  \bibinfo{person}{Tao Jiang}, {and} \bibinfo{person}{Xiaohong Guan}.}
  \bibinfo{year}{2021}\natexlab{}.
\newblock \showarticletitle{A review of deep reinforcement learning for smart
  building energy management}.
\newblock \bibinfo{journal}{\emph{IEEE Internet of Things Journal}}
  \bibinfo{volume}{8}, \bibinfo{number}{15} (\bibinfo{year}{2021}),
  \bibinfo{pages}{12046--12063}.
\newblock


\bibitem[Yu et~al\mbox{.}(2020)]%
        {yu2020multi}
\bibfield{author}{\bibinfo{person}{Liang Yu}, \bibinfo{person}{Yi Sun},
  \bibinfo{person}{Zhanbo Xu}, \bibinfo{person}{Chao Shen},
  \bibinfo{person}{Dong Yue}, \bibinfo{person}{Tao Jiang}, {and}
  \bibinfo{person}{Xiaohong Guan}.} \bibinfo{year}{2020}\natexlab{}.
\newblock \showarticletitle{Multi-agent deep reinforcement learning for HVAC
  control in commercial buildings}.
\newblock \bibinfo{journal}{\emph{IEEE Transactions on Smart Grid}}
  \bibinfo{volume}{12}, \bibinfo{number}{1} (\bibinfo{year}{2020}),
  \bibinfo{pages}{407--419}.
\newblock


\bibitem[Zhang et~al\mbox{.}(2019)]%
        {zhang2019building}
\bibfield{author}{\bibinfo{person}{Chi Zhang}, \bibinfo{person}{Sanmukh~R
  Kuppannagari}, \bibinfo{person}{Rajgopal Kannan}, {and}
  \bibinfo{person}{Viktor~K Prasanna}.} \bibinfo{year}{2019}\natexlab{}.
\newblock \showarticletitle{Building HVAC scheduling using reinforcement
  learning via neural network based model approximation}. In
  \bibinfo{booktitle}{\emph{Proceedings of the 6th ACM international conference
  on systems for energy-efficient buildings, cities, and transportation}}.
  \bibinfo{pages}{287--296}.
\newblock


\bibitem[Zhao et~al\mbox{.}(2021)]%
        {zhao2021hybrid}
\bibfield{author}{\bibinfo{person}{Huan Zhao}, \bibinfo{person}{Junhua Zhao},
  \bibinfo{person}{Ting Shu}, {and} \bibinfo{person}{Zibin Pan}.}
  \bibinfo{year}{2021}\natexlab{}.
\newblock \showarticletitle{Hybrid-model-based deep reinforcement learning for
  heating, ventilation, and air-conditioning control}.
\newblock \bibinfo{journal}{\emph{Frontiers in Energy Research}}
  \bibinfo{volume}{8} (\bibinfo{year}{2021}), \bibinfo{pages}{610518}.
\newblock


\bibitem[Zhao et~al\mbox{.}(2015)]%
        {zhao2015energyplus}
\bibfield{author}{\bibinfo{person}{Jie Zhao}, \bibinfo{person}{Khee~Poh Lam},
  \bibinfo{person}{B~Erik Ydstie}, {and} \bibinfo{person}{Omer~T Karaguzel}.}
  \bibinfo{year}{2015}\natexlab{}.
\newblock \showarticletitle{EnergyPlus model-based predictive control within
  design--build--operate energy information modelling infrastructure}.
\newblock \bibinfo{journal}{\emph{Journal of Building Performance Simulation}}
  \bibinfo{volume}{8}, \bibinfo{number}{3} (\bibinfo{year}{2015}),
  \bibinfo{pages}{121--134}.
\newblock


\bibitem[Zhuang et~al\mbox{.}(2023)]%
        {zhuang2023data}
\bibfield{author}{\bibinfo{person}{Dian Zhuang}, \bibinfo{person}{Vincent~JL
  Gan}, \bibinfo{person}{Zeynep~Duygu Tekler}, \bibinfo{person}{Adrian Chong},
  \bibinfo{person}{Shuai Tian}, {and} \bibinfo{person}{Xing Shi}.}
  \bibinfo{year}{2023}\natexlab{}.
\newblock \showarticletitle{Data-driven predictive control for smart HVAC
  system in IoT-integrated buildings with time-series forecasting and
  reinforcement learning}.
\newblock \bibinfo{journal}{\emph{Applied Energy}}  \bibinfo{volume}{338}
  (\bibinfo{year}{2023}), \bibinfo{pages}{120936}.
\newblock


\bibitem[Zou et~al\mbox{.}(2020)]%
        {zou2020towards}
\bibfield{author}{\bibinfo{person}{Zhengbo Zou}, \bibinfo{person}{Xinran Yu},
  {and} \bibinfo{person}{Semiha Ergan}.} \bibinfo{year}{2020}\natexlab{}.
\newblock \showarticletitle{Towards optimal control of air handling units using
  deep reinforcement learning and recurrent neural network}.
\newblock \bibinfo{journal}{\emph{Building and Environment}}
  \bibinfo{volume}{168} (\bibinfo{year}{2020}), \bibinfo{pages}{106535}.
\newblock


\end{thebibliography}

\appendix

\end{document}